\documentclass[letterpaper]{article} 
\usepackage[submission]{aaai25}  
\usepackage{times}  
\usepackage{helvet}  
\usepackage{courier}  
\usepackage[hyphens]{url}  
\usepackage{graphicx} 
\usepackage{natbib}  
\usepackage{caption} 
\usepackage{algorithm}
\usepackage{amssymb}
\usepackage{amsmath}
\usepackage{multirow}
\usepackage{algpseudocode}
\usepackage{xcolor}

\usepackage{newfloat}
\usepackage{listings}
\DeclareCaptionStyle{ruled}{labelfont=normalfont,labelsep=colon,strut=off} 
\floatstyle{ruled}
\newfloat{listing}{tb}{lst}{}
\floatname{listing}{Listing}
\title{Evidential Graph Contrastive Alignment for \\Source-Free Blending-Target Domain Adaptation}
\author{
    Juepeng Zheng, Yibin Wen, Jinxiao Zhang, Runmin Dong, Haohuan Fu 
}
\affiliations{
    Sun Yat-Sen University  \qquad    Tsinghua University

}

\usepackage{bibentry}

\begin{document}

\maketitle

\begin{abstract}
In this paper, we firstly tackle a more realistic Domain Adaptation (DA) setting: Source-Free Blending-Target Domain Adaptation (SF-BTDA), where we \textit{can not} access to source domain data while facing mixed multiple target domains without any domain labels in prior. 
Compared to existing DA scenarios, SF-BTDA generally faces
the co-existence of different label shifts in different targets, along with noisy target pseudo labels generated from the source model. In this paper, we propose a new method called Evidential Contrastive Alignment (ECA) to decouple the blending target domain and alleviate the effect from noisy target pseudo labels. First, to improve the quality of pseudo target labels, we propose a calibrated evidential learning module to iteratively improve both the accuracy and certainty of the resulting model and adaptively generate high-quality pseudo target labels. Second, we design a graph contrastive learning with the domain distance matrix and confidence-uncertainty criterion, to minimize the distribution gap of samples of a same class in the blended target domains, which alleviates the co-existence of different label shifts in blended targets. 
We conduct a new benchmark based on three standard DA datasets and ECA outperforms other methods with considerable gains and achieves comparable results compared with those that have domain labels or source data in prior.
\end{abstract}

%

\section{Introduction}
\label{sec:intro}

Despite the great success of deep learning in various fields \cite{krizhevsky2012imagenet,long2015fully,he2017mask}, it remains a challenge to achieve a generalized deep learning model that can perform well on unseen data, especially given the vast diversity of real-world data and problems. Performance degradation occurs due to various differences between the training data and new test data, for aspects such as statistical distribution, dimension, context, etc. 
Domain adaptation (DA) offers significant benefits in this scenario by adapting the pre-trained models towards new domains with different distributions and properties from the original domain \cite{ben2010theory,cui2020gradually}. Moreover, DA provides great potential for efficiency improvement by reducing the need for retraining on new data from scratch, which has been widely applied in various fields and real-world cases.







\begin{figure}[t]
    \centering  
    \includegraphics[width=0.40\textwidth]{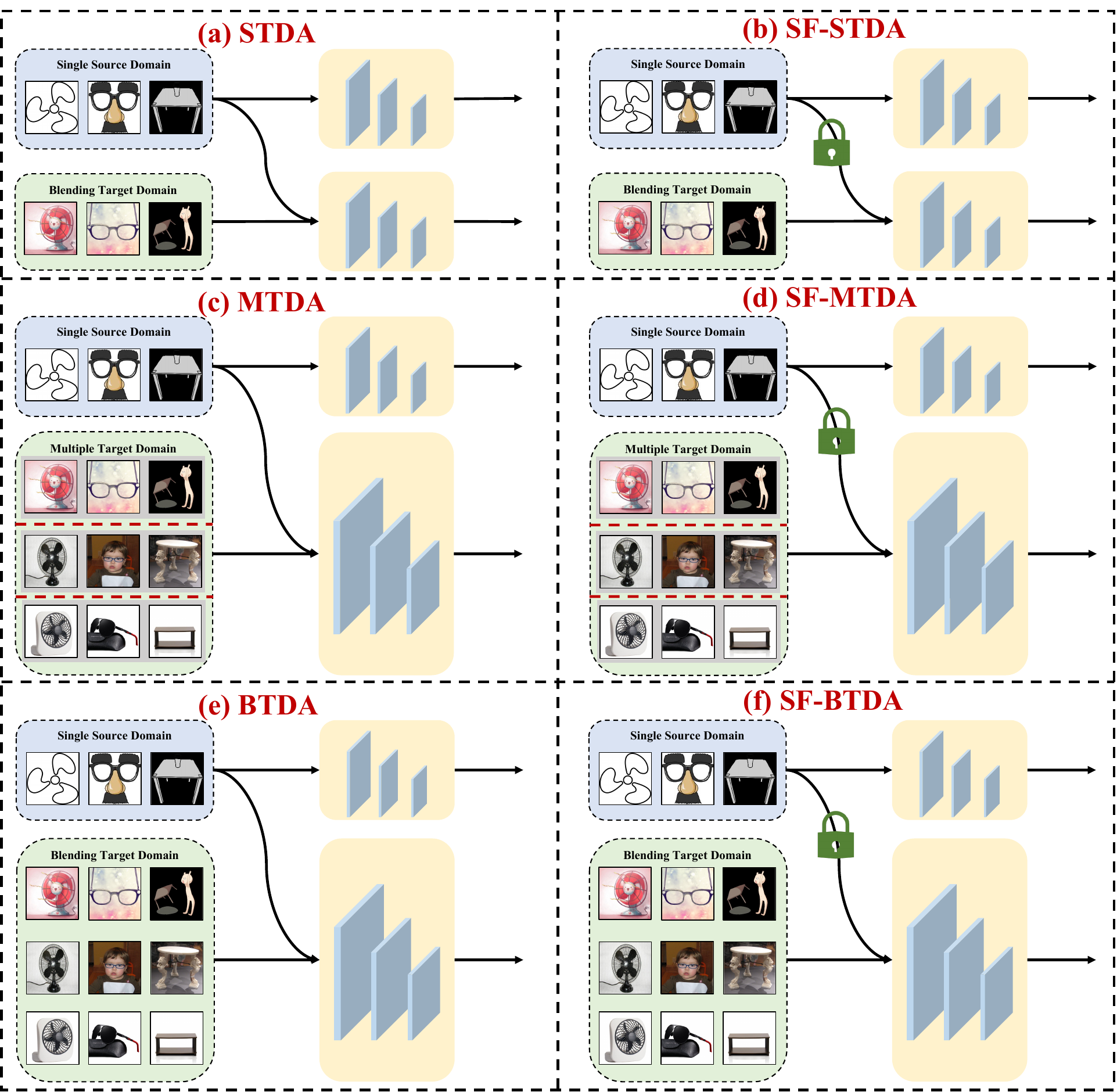}
    \caption{Different DA settings. (a) Single Target Domain Adaptation (STDA). (b) Source-Free STDA (SF-STDA). (c) Multiple Target Domain Adaptation (MTDA). (d) Source-Free MTDA (SF-MTDA). (e) Blending-Target Domain Adaptation (BTDA). (f) Source-Free BTDA (SF-BTDA), a new DA setting proposed in this paper, with only access to source trained model.}
    \label{fig:differentDA}
    \vspace{-2em}
\end{figure}

According to the number of target domains in the DA task, there are two main types of DA: Single-Target DA (STDA) (see Figure \ref{fig:differentDA}(a)) and Multi-Target DA (MTDA) (see Figure \ref{fig:differentDA}(c)). While most existing DA works focus on STDA \cite{ganin2016domain,long2016unsupervised,french2018self}, MTDA is a more challenging task that simultaneously transfer one source domain to multiple target domains. For example, MTDA-ITA \cite{gholami2020unsupervised}  aims to find a shared latent space common to all domains to allow adaptation from a single source to multiple target domains. HGAN \cite{yang2020heterogeneous}  aims to learn a unified subspace common for all domains with a heterogeneous graph attention network for better semantic transfer in MTDA. 

However, all aforementioned works have domain labels, which provides an important reference for deep learning models. However, real-world tasks are much more complex. For instance, in object recognition for autonomous driving technology, various changes may introduce new domains into the task, such as image collection device, street view, weather, lighting, time, road conditions and so on. Therefore, DA tasks in real-world applications often do not know their domain labels in prior, and the data from various target domains are mixed together. To our knowledge, there are only some works focusing on this mixed multi-target DA without domain labels, which is crucial for real-world applications \cite{chen2019blending,peng2019domain,roy2021curriculum}. Therefore, this paper makes efforts on the more practical and challenging Blending-Target Domain Adaptation (BTDA) scenario (see Figure \ref{fig:differentDA}(e)). 

On the other hand, there is an another crucial issue in real-world applications for DA, that is whether the source data is available. Owing to increasing privacy concerns (such as medical images, mobile applications, surveillance videos, etc.), source-domain data can not be always accessed. Recently, Source-Free Domain Adaptation (SFDA) is introduced to transfer knowledge from a prior source domain to a new target domain without accessing the source data. As shown in Figure \ref{fig:differentDA}, we could define different SFDA settings according to the number of target domains in the DA task. However, most existing SFDA methods focus on SF-STDA setting (see Figure \ref{fig:differentDA}(b)) \cite{kundu2020universal,liu2021source,zhang2023class}, while only \cite{kumar2023conmix} have discussed and designed new approach for SF-MTDA (see Figure \ref{fig:differentDA}(d)). 

Different from other DA scenarios, we proposed a new DA setting named Source-Free Blending-Target Domain Adaptation (SF-BTDA), where we have to transfer the knowledge from a single source to blending-target domain without domain labels nor the access of labeled source data (see Figure \ref{fig:differentDA}(f)). There are two major challenges in SF-BTDA: \textbf{\textit{Challenge 1}}: facing a mixture of multiple target domains, the model is assumed to handle the co-existence of different label shifts in different targets with no prior information;
\textbf{\textit{Challenge 2}}: as we are only accessed to the source model, precisely alleviating the negative effect from noisy target pseudo labels is difficult because of the significantly enriched styles and textures. Due to these two challenges, existing DA methods that count on utilizing the distribution shift between two specific domains become infeasible.

In this paper, we first present SF-BTDA setting and propose Evidential Contrastive Alignment (ECA) to better tackle the aforementioned two challenges in SF-BTDA scenarios. The contributions of this work are as following:




\begin{itemize}
    \item We propose a novel DA task named SF-BTDA, where we have to transfer the knowledge from a single source to blending-target domain without domain labels nor the access of labeled source data.
    \item  We propose calibrated evidential learning to iteratively improve both the accuracy and certainty of the resulting model and adaptively select high-quality pseudo target labels according to the balanced accuracy and certainty.
    \item We design graph contrastive learning with the domain distance matrix and confidence-uncertainty criterion, to minimize the distribution gap of samples of a same class in the blended-target domain.
    \item We conduct a new benchmark for SF-BTDA setting and comprehensive experiments show that ECA outperforms other methods with considerable gains and achieves comparable results compared with those that have domain labels or source data in prior.
    
\end{itemize}

The second contribution mainly address the \textbf{\textit{Challenge 2}}, improving the quality of pseudo target labels in the blending target domain to reduce the effect of the enriched styles and textures in the mixture of multiple target domains.
The third contribution effectively tackles the \textbf{\textit{Challenge 1}}, forcefully pulling samples that are from the same class to alleviate the co-existence of different label shifts in blended targets.

\section{Related works}
\label{sec:related}

\subsection{Domain Adaptation}
Domain adaptation (DA) aims to minimize domain shift, which refers to the distribution difference between the source and target domains.  Most of them focus on STDA. Plenty of single target DA methods have been devised to different tasks including classification \cite{ganin2016domain}, semantic segmentation \cite{li2019bidirectional} and object detection \cite{hsu2020progressive}. 
To reduce domain discrepancies, there are different kinds of methods. One popular approach is motivated by generative adversarial networks (GAN) \cite{goodfellow2014generative}. Adversarial networks are widely explored and try to minimize the domain shift by maximizing the confusion between source and target \cite{xia2021adaptive}. Alternatively, it is also common practice to adopt statistical measure matching methods to bridge the difference between source and target, including MMD \cite{long2015learning}, JMMD \cite{long2017deep}, and other variation metrics \cite{kang2019contrastive}.
Although extensive works have been done on STDA, there is less research on multi-target DA, especially for blending-target DA, which is more similar to the real-world scenarios.

\subsection{MTDA and BTDA}
Different from STDA, MTDA is more flexible because the model can be adapted to multiple target domains with different characteristics, which requires the model to generalize better on target data without annotations \cite{gholami2020unsupervised,nguyen2021unsupervised}. These requirements introduce new challenges, especially the computational requirements and catastrophic forgetting of previously-learned targets. 
However, BTDA has received little attention. This approach aims to provide solutions for situations where there are no explicit domain labels and samples from different target domains are mixed together, which is more consistent with reality. 
The differences between single-target DA, multi-target DA and mixed multi-target DA is shown in Fig. \ref{fig:differentDA}. To the best of our knowledge, there are only few works related to this field. For instance, DADA \cite{peng2019domain} aims to tackle domain-agnostic learning by disentangling the domain-invariant features from both domain-specific and class-irrelevant features simultaneously. CGCT \cite{roy2021curriculum} uses curriculumn learning and co-teaching methods to obtain more reliable pseudo-labels.  
In addition, Open Compound Domain Adaptation (OCDA) also contains blending target domain \cite{liu2020open,park2020discover,gong2021cluster}, while it contains extra "unseen" domain, which is beyond the scope of this paper.

\subsection{Souce-Free Domain Adaptation}
Because of the limit of privacy constraints, SFDA is becoming more and more popular, which only provides the well-trained source model without any access to source data during the adaptation process. There are two mainstreams for SFDA methods. On the one hand, some methods focus on reconstructing the pseudo source domain samples in the feature space \cite{qiu2021source,tian2021vdm}
On the other hand, some methods exploit pseudo target labels from source model and adopt self-training or noisy learning so that the model is well-fitted to the target domain \cite{chen2022self}. For example, SHOT \cite{liang2020we} adopts entropy minimization and information maximization through pseudo-labeling learning to transfer the knowledge from trained classifier to target features. 
Up to date, Most Existing SFDA methods focus on SF-STDA setting, while only \cite{kumar2023conmix} have discussed and designed new approach for SF-MTDA. ConMix \cite{kumar2023conmix} introduces consistency between label preserving augmentations and utilizes pseudo label refinement methods to reduce noisy pseudo labels for both SF-STDA and SF-MTDA settings.

However, existing literature have not exploited a new DA setting named Source-Free Blending-Target Domain Adaptation (SF-BTDA), where we have to transfer the knowledge from a single source to blending-target domain without domain labels nor the access of labeled source data. 
In this paper, we propose Evidential Contrastive Alignment (ECA) to better address both two major challenges for SF-BTDA.

\section{Methodology}
\label{sec:method}


\begin{figure*}[t]
    \centering  
    \includegraphics[width=0.95\textwidth]{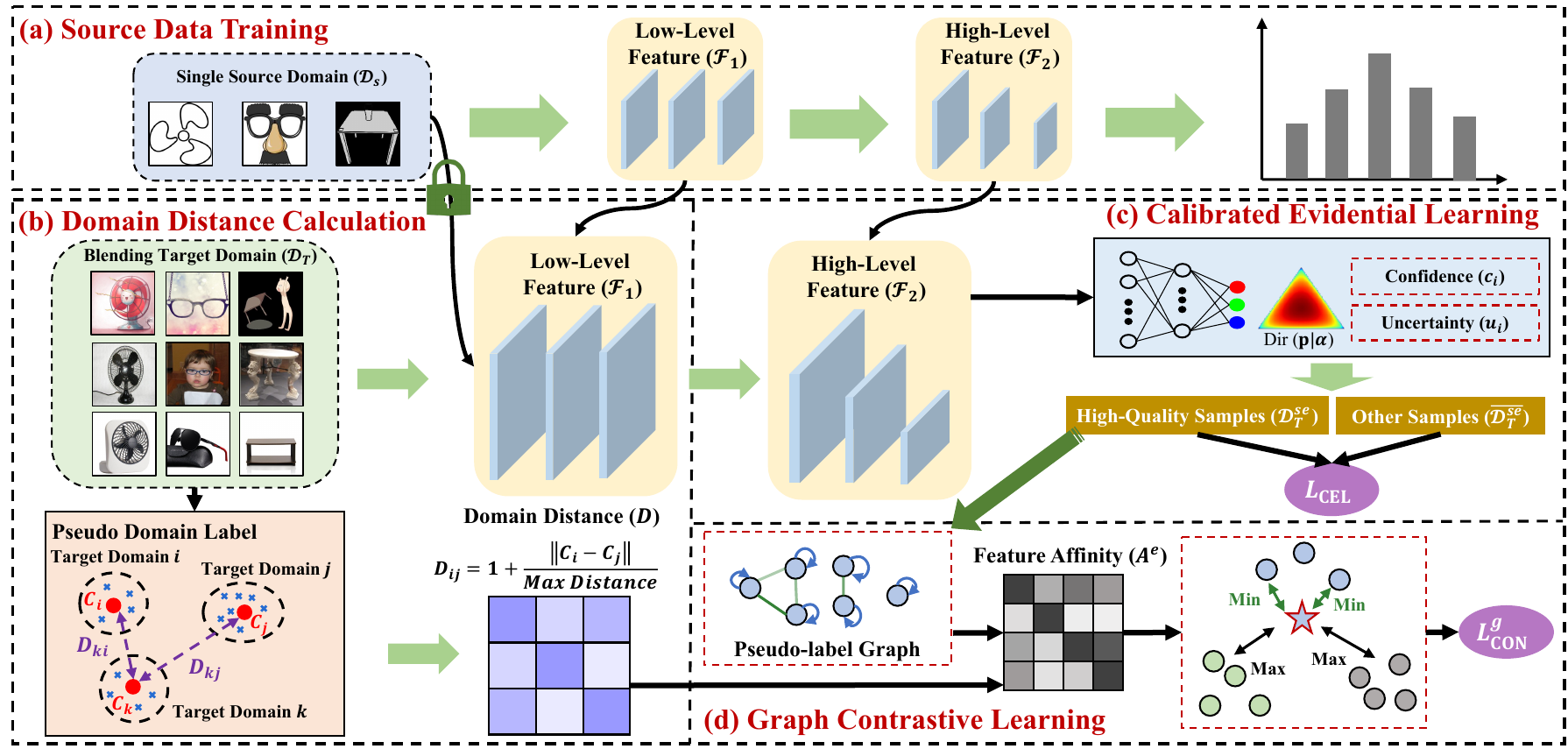}
    \caption{The framework for our proposed Evidential Contrastive Alignment (ECA). (a) \textbf{Source Data Training}: We train the model through single source domain and we are not access to the source domain data. (b) \textbf{Domain Distance Calculation}: We generate the mixture domain distance through low-level features and original image textures in an unsupervised way. (c) \textbf{Calibrated Evidential Learning}: We propose a calibrated evidential learning module to iteratively improve both the accuracy and certainty of the resulting model and adaptively generate high-quality pseudo target label (addressing \textbf{\textit{Challenge 1}}). (d) \textbf{Graph Contrastive Learning}: We design a graph contrastive learning with the domain distance matrix and confidence-uncertainty criterion, to minimize the distribution gap of samples of a same class in the blended target domains, which alleviates the co-existence of different label shifts in blended targets (addressing \textbf{\textit{Challenge 2}}).}
    \label{fig:framework}
    \vspace{-2em}
\end{figure*}

\subsection{Problem Setting and Notations}
In contrast to MTDA, BTDA is established on a mixture of target domains $\mathcal{D}_T=\left\{\mathcal{D}_t^j \right\}_{j=1}^k=\left\{(\mathbf{x}_i^t )\right\}_{i=1}^{n_T}$, where $k$ is the number of target domains and $n_T=\sum_{j=1}^kn_t^j$  denotes the total quantity of images in the target domain. Unlike the MTDA scenario, the proportions of different target domains in the mixed datasets $\left\{w^j \right\}_{j=1}^k$ are unknown. 
In the SF-BTDA setting, we consider the labeled source domain data with ${n_s}$, the samples as $\mathcal{D}_s=\left\{\left(\mathbf{x}_i^s, \mathbf{y}_i^s \right)\right\}_{i=1}^{n_s}$, in which the $\mathbf{y}_i^s$ is the corresponding label of $\mathbf{x}_i^s$, is only available during model pretraining. Our proposed approach is based on any backbone, 
which we split into three parts: a shallow feature extractor $\mathcal{F}_1$, a deep feature extractor $\mathcal{F}_2$ and a classifier $\mathcal{G}$.

As we have to transfer the knowledge from a single source to blending-target domain without domain labels nor the access of labeled source data, if we directly adopt existing SF-STDA algorithms and consider the mixed target domains as one target domain in a brute-force way, the training objective will facilitate domain-invariant representations to align the whole blending-target domain $\mathcal{D}_T$ rather than $k$ target domains $\left\{\mathcal{D}_t^j \right\}_{j=1}^k$. Because of the discrepancy among sub-subjects from $k$ distributions, adaptation for the whole blending-target domain directly may  lead to severe category misalignment and  negative transfer effects. 


\subsection{Domain Distance Calculation}
\label{sec:DDC}
As introduced in Sec. \ref{sec:intro}, SF-BTDA issue faces a mixture of multiple target domains, the model is assumed to handle the co-existence of different label shifts in different targets with no prior information (\textbf{\textit{Challenge 1}}). Therefore, some samples in the same class may appear different styles and textures because of co-existence multiple targets with significantly enriched styles and textures. 
Under this consideration, we calculate the domain distance to better guide the contrastive learning (see Sec. \ref{sec:CFA}) to pull the samples  in the same class that may appear different styles and textures.
As we are not access to the specific domain label for the blending-target domain, we separate the target domains into $k$ domains and assign pseudo domain labels for them using an unsupervised way (\textit{i.e.}, $k$-means). Notably, we adopt the extracted shallow features from DNNs ($\mathcal{F}_1\left(\mathbf{x}^t\right)$) and integrate them with original image textures ($\mathbf{x}^t$) to achieve pseudo domain assignment for better satisfying the specific image style as well as the universal feature style. Notably, we simply concatenate  the low-level feature with the original image $x_i$ and we adopt the low-level features that generated from conv2\_x with the dimension of 56$\times$56$\times$256.

We calculate the domain distance $D_{ij}$ between two samples $\mathbf{x}^t_i$ and $\mathbf{x}^t_j$ through Eq. \eqref{eq:domain_dis}, in which $C_i$ and $C_j$ are the corresponding centroids for $\mathbf{x}^t_i$ and $\mathbf{x}^t_j$. The centroid for an target image ($x^t_i$) denotes it belongs to the pseudo domain label. The $Max \ Distance$ denotes we normalize the domain distance among different target domain centroids. If $\mathbf{x}^t_i$ and $\mathbf{x}^t_j$ are from the same domain, the domain distance will be $1.0$. In this paper, we utilize the distance of domain centroids to represent the samples from different target domains for simplicity. Notably, we update the domain distance after each training epoch. 

\begin{equation}
    D_{ij} = 1 + \frac{\Vert C_i - C_j \Vert}{Max \ Distance} 
    \label{eq:domain_dis}
\end{equation}

%


\subsection{Calibrated Evidential Learning}
\label{sec:CEU}

In the meantime, as we are also not access to the source data, we could only achieve pseudo target labels for the blending-target images using the model trained from labeled source images. Because of \textbf{\textit{Challenge 2}} introduced in Sec. \ref{sec:intro}, directly adopting pseudo labels for the target domain may ruin and mislead the model training. To alleviate the negative effect  from noisy target labels, we propose calibrated evidential learning to  to iteratively improve both the accuracy and certainty of the resulting model, and adaptively select high-quality pseudo target labels. 

Evidential Deep Learning (EDL) \cite{amini2020deep,bao2021evidential} offers a principled way to formulate the uncertainty estimation, which overcomes the shortcoming of softmax-based DNNs. Assuming that the class probability follows a prior Dirichlet distribution \cite{sentz2002combination}, we can acquire the predictive uncertainty $u$ for each sample: $u = M / S$, in which $M$ is the number of classes, and  $S$ is the total evidence that can be defined as $S=\sum_{m=1}^M \mathbf{\alpha}_m$. $\mathbf{\alpha}_m$ is the non-negative network prediction output and can be expressed $\mathbf{\alpha}_m = \mathbf{e}_m + 1$, where $\mathbf{e}(i)=g(\mathcal{F}_2(\mathcal{F}_1(\mathbf{x}(i))))$ and $g$ is the evidence function. More detailed information could be seen in supplementary. 



%

Although the evidential uncertainty from EDL can be directly learned in the target domain samples, the uncertainty may not be well calibrated to address our target pseudo label learning in SF-BTDA scenario. According to existing model calibration literature \cite{krishnan2020improving}, a better calibrated model is supposed to be uncertain in its predictions when being inaccurate, while be confident about accurate ones.  
In this section, we propose to calibrate the EDL model by considering the relationship between the accuracy and uncertainty for the blending-target domain under source-free setting. 

In particular, we propose Calibrated Evidential Learning (CEL) method to minimize the following sum of samples that are accurate but uncertainty, or inaccurate but certain, using the logarithm constraint between the confidence $c_i$ and uncertainty $u_i$:

\vspace{-1em}
\begin{equation}
\begin{aligned}
    L_{\text{CEL}} &= 
    -\lambda_t \sum_{\mathbf{x}_i \in \mathcal{D}_T^{\text{se}}} c_i \text{log}(1 - u_i) \\
            &-(1 - \lambda_t) \sum_{\mathbf{x}_i \in \overline{\mathcal{D}_T^{\text{se}}}} (1 - c_i) \text{log}(u_i)
\end{aligned}
\label{eq:evidential}
\end{equation}
\vspace{-1em}

in which $u_i$ denotes the evidential uncertainty of an input sample $\mathbf{x}_i$ and $c_i$ is the associated maximum class probability. $\mathcal{D}_T^{\text{se}}$ and $\overline{\mathcal{D}_T^{\text{se}}}$ are the selected high-quality target samples and the remainder in the target domain (low-quality target samples), respectively. That is, $|\mathcal{D}_T| = |\mathcal{D}_T^{\text{se}}| + |\overline{\mathcal{D}_T^{\text{se}}}|$. 

Different from other high-quality selection schemes in SFDA setting \cite{guillory2021predicting,dong2021confident,wang2021uncertainty,karim2023c}, we utilize the evidential uncertainty and prediction confidence to measure the label reliability simultaneously. That is, the high-quality target pseudo labels both satisfy  that $c_i > \eta_c$ and $u_i > \eta_u$, where $\eta_c$ and $\eta_u$ are two selection thresholds that can be adaptively estimated in each mini-batch. To this end, we do not need to set fixed hyper-parameters for $\eta_c$ and $\eta_u$ (More comparisons can be found in supplementary).

\vspace{-1em}
\begin{equation}
    \eta_c = \frac{1}{B} \sum_{i=1}^{i=B} c_i; \ \ \ \ \ \eta_u = \frac{1}{B} \sum_{i=1}^{i=B} u_i
    \label{eq:selection}
\end{equation}

The first term of $L_{\text{CEL}}$  in Eq. \eqref{eq:evidential} tries to give low uncertainty ($u_i \rightarrow 0$) when the model makes accurate prediction ($\mathbf{x}_i \in \mathcal{D}_T^{\text{se}}, c_i \rightarrow 1$) for the selected high-quality target samples, while the second term of $L_{\text{CEL}}$ aims to give high uncertainty ($u_i \rightarrow 1$) when the model makes inaccurate prediction ($\mathbf{x}_i \in \overline{\mathcal{D}_T^{\text{se}}}, c_i \rightarrow 0$) for the low-quality target samples. Notably, $\gamma_t$ is the annealing factor that will be exponentially increasing from $\gamma_0$ to $1.0$ during training process (More details can be found in our supplementary). 


To this end, through our proposed calibrated evidential learning, the model tries to make high accuracy and low uncertainty for the high-quality target domain samples, while make a restraint for low-quality target domain samples, which addresses the \textbf{\textit{Challenge 2}} introduced in Sec. \ref{sec:intro}. This strategy helps us to better achieve the knowledge transfer from the  source model to the blending-target domain. 


\subsection{Graph Contrastive Learning}
\label{sec:CFA}




The original intention for contrastive learning is to push each image far from others and pull the augmented images closer with the original one in the feature space. 
However, it is not compatible with the classification task which requires to cluster images at the class level, especially that we have some high-quality target samples generated from Sec. \ref{sec:CEU}. Furthermore, according to the description of \textbf{\textit{Challenge 1}}, some samples in the same class may appear different styles and textures because of co-existence multiple targets with significantly enriched styles and textures. 
To this end, we design the domain distance calculated from Sec. \ref{sec:DDC} in graph contrastive learning to pay more attention for abovementioned samples, which could solve the conflict between contrastive learning and the classification task.

\begin{table*}[t]
\centering
\small
\begin{tabular}{cccc|cccc|ccccc}
\hline
\multirow{2}{*}{\textbf{Method}} &  \textbf{{Multiple}} & \textbf{{Domain}} & \textbf{{Source}} & \multicolumn{4}{c}{\textbf{Office-31}} & \multicolumn{5}{c}{\textbf{OfficeHome}} \\ 
 & \textbf{Targets} & \textbf{Labels} & \textbf{Data} & \textbf{A}    & \textbf{D}  &  \textbf{W} & \textbf{Avg} & \textbf{Ar}    & \textbf{Cl}  &  \textbf{Pr} & \textbf{Rw} &  \textbf{Avg}  \\ \hline
ResNet-50 \shortcite{he2016deep} & $\times$ & $\times$ & $\times$ & 76.3 & 68.7 & 67.0 & 70.7 & 62.5 & 61.2 & 55.1 & 61.8 & 60.1 \\ \hline
DAN \shortcite{long2015learning} & $\times$ & $\times$ & $\checkmark$ & 79.5 & 80.3 & 81.2 & 80.3 & 58.4 & 58.1 & 52.9 & 62.1 & 57.9 \\ 
DANN \shortcite{ganin2016domain} & $\times$ & $\times$ & $\checkmark$ & 78.2 & 72.2 & 69.8 & 73.4 & 58.4 & 58.1 & 52.9 & 62.1 & 57.9 \\
CDAN \shortcite{long2018conditional} & $\times$ & $\times$ & $\checkmark$ & 93.6 & 80.5 & 81.3 & 85.1 & 59.5 & 61.0 & 54.7 & 62.9 & 59.5\\\hline
MTDA-ITA \shortcite{gholami2020unsupervised} & $\checkmark$ & $\checkmark$ & $\checkmark$ & 87.9 & 83.7 & 84.0 & 85.2 & 64.6 & 66.4 & 59.2 & 67.1 & 64.3 \\
DCL \shortcite{nguyen2021unsupervised} & $\checkmark$ & $\checkmark$ & $\checkmark$ & 92.6 & 82.5 & 84.7 & 86.6 & 63.0 & 63.0 & 60.0 & 67.0 & 63.3 \\
DCGCT \shortcite{roy2021curriculum} & $\checkmark$ & $\checkmark$ & $\checkmark$ & 93.4 & 86.0 & 87.1 & 88.8 & 70.5 & 71.6 & 66.0 & 71.2 & 69.8 \\ \hline
AMEAN \shortcite{chen2019blending} & $\checkmark$ & $\times$ & $\checkmark$ & 90.1 & 77.0 & 73.4 & 80.2 & 64.3 & 65.5 & 59.5 & 66.7 & 64.0\\
CGCT \shortcite{roy2021curriculum} & $\checkmark$ & $\times$ & $\checkmark$ & 93.9 & 85.1 & 85.6 & 88.2 & 67.4 & 68.1 & 61.6 & 68.7 & 66.5\\
DML \shortcite{wang2022discriminative} & $\checkmark$ & $\times$ & $\checkmark$ & \underline{94.9} & 85.3 & 86.2 & 88.8 & 67.0 & 70.6 & 62.9 & 69.0 & 67.4\\
MCDA \shortcite{xu2023class} & $\checkmark$ & $\times$ & $\checkmark$ & 92.4 & \underline{87.7} & \underline{88.8} & \underline{89.6}  & {71.7} & 72.8 & {68.0} & {71.7} & {71.1} \\ \hline
G-SFDA \shortcite{yang2021generalized} & $\times$ & $\times$ & $\times$ & 90.6 & 78.9 & 77.4 & 82.3 & 70.3 & {74.9} & 66.8 & 70.0 & 70.5 \\
GPUE \shortcite{litrico2023guiding} & $\times$ & $\times$ & $\times$ & 78.8 & 72.4 & 60.7 & 70.6 & 50.5 & 45.5 & 46.4 & 52.9 & 48.8 \\
SHOT++ \shortcite{liang2021source} & $\times$ & $\times$ & $\times$ & {{93.3}} & 79.6 & 78.4 & 83.8 & {70.6} & {74.2} & 66.2 & 68.6 & 69.9 \\ \hline
{ConMix \shortcite{kumar2023conmix}}  & $\checkmark$ & $\checkmark$ & $\times$ & 92.4 & 81.8 & {80.4} & 84.9 & {75.6} & {81.4} & 71.4 & 73.4 & {75.4} \\ \hline
ECA (ours) & $\checkmark$ & $\times$ & $\times$ & \textbf{93.5} & {\textbf{84.4}} & \textbf{81.3} & {\textbf{86.4}} & \underline{\textbf{76.5}} & \underline{\textbf{82.2}} & {\underline{\textbf{72.2}}} & {\underline{\textbf{73.6}}} & \underline{\textbf{76.1}} \\\hline

\end{tabular}
\caption{Overall Accuracy (\%) on \textbf{Office-31} and \textbf{OfficeHome} for for state-of-the-art methods. All methods use the ResNet-50 as the backbone.  The results on each dataset are the average of  three/four leave-one-domain-out cases.}
\label{tab:office}
\vspace{-1em}
\end{table*}

\begin{table*}[t]
\centering
\small
\begin{tabular}{cccc|ccccccc}
\hline
\textbf{\multirow{2}*{Method}} &  \textbf{{Multiple }} & \textbf{{Domain}} & \textbf{{Source}}  & \textbf{\multirow{2}*{Cli}}    & \textbf{\multirow{2}*{Inf}}  &  \textbf{\multirow{2}*{Pai}} & \textbf{\multirow{2}*{Qui}} & \textbf{\multirow{2}*{Rea}}    & \textbf{\multirow{2}*{Ske}}  &  \textbf{\multirow{2}*{Avg}}  \\
& \textbf{Targets} & \textbf{Labels} & \textbf{Data} \\  \hline
ResNet-101 \shortcite{he2016deep} & $\times$ & $\times$ & $\times$ & 25.6 & 16.8 & 25.8 & 9.2 & 20.6 & 22.3 & 20.1 \\ \hline
SE \shortcite{french2018self} & $\times$ & $\times$ & $\checkmark$ & 21.3 & 8.5 & 14.5 & 13.8 & 16.0 & 19.7 & 15.6 \\ 
MCD \shortcite{saito2018maximum} & $\times$ & $\times$ & $\checkmark$ & 25.1 & 19.1 & 27.0 & 10.4 & 20.2 & 22.5 & 20.7 \\
CDAN \shortcite{long2018conditional} & $\times$ & $\times$ & $\checkmark$ & 31.6 & 27.1 & 31.8 & 12.5 & 33.2 & 35.8 & 28.7 \\
MCC \shortcite{jin2020minimum} & $\times$ & $\times$ & $\checkmark$ & 33.6 & 30.0 & 32.4 & 13.5 & 28.0 & 35.3 & 28.8 \\\hline
DCL \shortcite{nguyen2021unsupervised} & $\checkmark$ & $\checkmark$ & $\checkmark$ & 35.1 & 31.4 & 37.0 & 20.5 & 35.4 & 41.0 & 33.4 \\
DCGCT \shortcite{roy2021curriculum} & $\checkmark$ & $\checkmark$ & $\checkmark$ & 37.0 & 32.2 & 37.3 & 19.3 & 39.8 & 40.8 & 34.4  \\ \hline
DADA \shortcite{peng2019domain} & $\checkmark$ & $\times$ & $\checkmark$ & 26.4 & 20.0 & 26.5 & 12.9 & 20.7 & 22.8 & 21.6\\ 
CGCT \shortcite{roy2021curriculum} & $\checkmark$ & $\times$ & $\checkmark$ & 36.1 & 33.3 & 35.0 & 10.0 & \underline{39.6} & 39.7 & 32.3 \\
DML \shortcite{wang2022discriminative} & $\checkmark$ & $\times$ & $\checkmark$ & 32.0 & 25.4 & 29.4 & 12.7 & 31.5 & 36.4 & 27.9 \\
MCDA \shortcite{xu2023class} & $\checkmark$ & $\times$ & $\checkmark$ & 37.5 & \underline{37.3} & 36.6 & {17.8} & 36.1 & {41.4} & {34.5} \\ \hline
G-SFDA \shortcite{yang2021generalized} & $\times$ & $\times$ & $\times$ & 34.2 & 26.7 & 32.7 & 11.7 & 18.7 & 32.0 & 26.0 \\
GPUE \shortcite{litrico2023guiding} & $\times$ & $\times$ & $\times$ & 23.4 & 22.7 & 31.4 & 16.5 & 16.9 & 24.8 & 22.6 \\
SHOT++ \shortcite{liang2021source} & $\times$ & $\times$ & $\times$ & 32.9 & 28.8 & 32.1 & 13.9 & 17.9 & 34.9 & 26.7  \\ \hline
{ConMix \shortcite{kumar2023conmix}}  & $\checkmark$ & $\checkmark$ & $\times$ & {41.8} & 29.2 & {39.9} & {17.5} & 32.7 & {41.2} & 33.7 \\  \hline
ECA (ours) & $\checkmark$ & $\times$ & $\times$ & \underline{\textbf{42.2}} & {\textbf{32.6}} & \underline{\textbf{40.3}} & \underline{\textbf{18.0}} & {\textbf{35.5}} & \underline{\textbf{41.8}} & \underline{\textbf{34.6}} \\\hline

\end{tabular}
\caption{Overall Accuracy (\%) on \textbf{DomainNet} for for state-of-the-art methods. All methods use the ResNet-101 as the backbone.  The results on each dataset are the average of six leave-one-domain-out cases.}
\vspace{-2em}
\label{tab:domainnet}

\end{table*}

Since Sec. \ref{sec:CEU} has generated high-quality labels for the target domain, we could fully explored their advantages in the contrastive learning. 
Notably, our contrastive alignment module aims to learn representations guided by a pseudo-label graph. Here we build the pseudo-label graph by constructing a similarity matrix ($A$). 
To this end, each element $a_{ij}$ in $A$ has the following format: 

\vspace{-1em}
	\begin{eqnarray}
		a_{ij}=\left\{
		\begin{aligned}
			1 & \quad \text{if} \ j = i, \\
			1 & \quad \text{if} \ z_i \  \text{and} \  z_j \  \text{are  from  same  category},  \\
                & \quad  \text{and} \  \mathbf{x}_i \in {\mathcal{D}^{se}_T} \  \text{and} \  \mathbf{x}_j \in {\mathcal{D}^{se}_T}, \\
			0 & \quad \text{otherwise} 
		\end{aligned}
		\right.
	\end{eqnarray}

Notably $\mathbf{x}_i \in {\mathcal{D}_T^{se}}$   and $\mathbf{x}_j \in {\mathcal{D}_T^{se}}$ denotes that $\mathbf{x}_i$ and $\mathbf{x}_j$ are selected from the high-quality target domain samples.
Although high-quality pseudo label-based supervised contrastive learning is theoretically functional, directly using these pseudo labels may be problematic because they might be noisy labels, too. To alliviate this problem, we adopt the confidence-uncertainty evaluation produced from Sec. \ref{sec:CEU} to strengthen the robustness and reduce the negative effect on noisy pseudo labels. Actually, the pseudo-label graph ($A$) serves as the target to train an embedding graph ($A^e$), which is defined as:  

\vspace{-1em}
\begin{eqnarray}
		a_{ij}^e=\left\{
		\begin{aligned}
			 a_{ij} * I_{ij} *D_{ij}  &\ \ \ \ \  \text{if} \ i \neq j, \\
			a_{ij} &\ \ \ \ \   \text{otherwise} 
		\end{aligned}
		\right.
\end{eqnarray}

in which $I_{ij} = c_i * c_j * (1 - u_i) * (1 - u_j)$ denotes that samples with high confidence and low uncertainty will have higher weights in our evidential contrastive alignment module. $D_{ij}$ can be calculated from Eq. \eqref{eq:domain_dis}, which could make our contrastive learning module to pay more attention on the samples of a same class that belonged to different target domains (addressing \textbf{\textit{Challenge 1}}). 
To this end, the loss of $L_{\text{CON}}^e$ can be formulated as:
 \begin{equation}
 \begin{aligned}
     L_{\text{CON}}^e =  &-\sum_{i \in \mathcal{I}} \frac{1}{1 + |\mathcal{P}(i)|} \{\text{log} \frac{\text{exp}(z_i \cdot z_{j(i)} / \tau)}{\sum_{q \in Q(i)} \text{exp}(z_i \cdot z_q / \tau)} \\
     &-\sum_{p \in \mathcal{P}(i)} \text{log}\frac{a_{ij}^e \cdot \text{exp}(z_i \cdot z_p / \tau)}{\sum_{q \in Q(i)} \text{exp}(z_i \cdot z_q / \tau)}   \}
 \end{aligned}
\end{equation}

in which $\mathcal{P}(i)$ denotes indices of the views from other images of the same class in the selected high-quality target domain samples. $|\mathcal{P}(i)|$ denotes its number and $|\mathcal{P}(i)| + 1$ denotes all positive pairs. $\tau$ is the temperature parameter. Other notations could refer to \cite{khosla2020supervised}. As our setting is SF-BTDA, it is noteworthy that our graph contrastive learning module conducts on the blending target owing to its difficulties from classification confusions because of co-existence of different label shifts (addressing \textbf{\textit{Challenge 1}}).

\subsection{Overall Objective}

The model is jointly optimized with two loss terms, including  calibrated evidential learning loss $L_{\text{CEL}}$ and graph contrastive learning loss $L_{\text{CON}}^e$:

\begin{equation}
    L = L_{\text{CEL}} + \beta L_{\text{CON}}^e
    \label{eq:total}
\end{equation}

where $\beta$ is a trade-off parameter balancing two different loss terms between $L_{\text{CEL}}$, and $L_{\text{CON}}^e$. 


\section{Experiments}

\subsection{Datasets}

We conduct experiments on three standard domain adaptation benchmarks. \textbf{Office-31} \cite{saenko2010adapting} consists of 4,652 images from three domains: DSLR (D), Amazon (A), and Webcam (W). \textbf{OfficeHome} \cite{venkateswara2017deep} is a more challenging dataset, which consists of
15,500 images in total from 65 categories of everyday objects. There are four significantly different domain: Artistic images (Ar), Clip-Art images (Cl), Product images (Pr), and Real-World images (Rw).  \textbf{DomainNet} \cite{peng2019moment} is the most challenging and very large scale DA benchmark, which has six different domains: Clipart (Cli), Infograph (Inf), Painting (Pai), Quickdraw (Qui), Real (Rea) and Sketch (Ske). It has around 0.6 million images, including both train and test images, and has 345 different object categories.

\subsection{Training Details}

Our methods were implemented based on the PyTorch \cite{paszke2019pytorch}. We adopt ResNet-50 \cite{he2016deep} for \textbf{Office-31} and \textbf{OfficeHome}, and ResNet-101 for \textbf{DomainNet}, both of which  pretrained on the ImageNet dataset \cite{russakovsky2015imagenet}. Whatever module trained from scratch, its
learning rate was set to be 10 times that of the lower layers. We adopt mini-batch stochastic gradient descent (SGD) with momentum of 0.95 using the learning rate and progressive training strategies. We set $\beta$ in Eq \eqref{eq:total} as 1.0. 
Our codes are available on \url{https://github/}

\textbf{State-of-the-art.} ResNet \cite{he2016deep} is the baseline backbone without any domain adaptation tricks. We list some methods that proposed for STDA scenario, such as DAN \cite{long2015learning}, DANN \cite{ganin2016domain}, RTN \cite{long2016unsupervised}, JAN \cite{long2017deep}, SE \cite{french2018self}, MCD \cite{saito2018maximum}, CDAN \cite{long2018conditional} and MCC \cite{jin2020minimum}. We also compare our method with existing three MTDA methods (\textit{i.e.}, MTDA-ITA \cite{gholami2020unsupervised}, DCL \cite{nguyen2021unsupervised}, DCGCT \cite{roy2021curriculum}) and five BTDA methods (\textit{i.e.}, AMEAN \cite{chen2019blending}, DADA \cite{peng2019domain}, CGCT \cite{roy2021curriculum}, DML \cite{wang2022discriminative} and MCDA \cite{xu2023class}). Furthermore, we compare three source-free DA methods, including G-SFDA \cite{yang2021generalized}, GPUE \cite{litrico2023guiding} and SHOT++ \cite{liang2021source}. However, these SFDA methods mainly focus on SF-STDA settings.
Finally, we compare our performance with CoNMix \cite{kumar2023conmix}, which is the only one that concentrates on SF-MTDA scenarios with domain labels in prior while without the access to the labels of source data.



\subsection{Results}

The \textbf{bold} numbers in Table \ref{tab:office} and \ref{tab:domainnet} denote the highest accuracy without source data. The \underline{underline} numbers denote the highest accuracy without domain labels. 
According to Table \ref{tab:office} and \ref{tab:domainnet}, our ECA overpasses these comparison methods without domain labels nor source data on all transfer tasks. Notably, MCDA and CGCT must use source data in prior. However, our ECA \textit{does not} use the source data nor domain labels. Therefore, it is surprising that ECA outperforms these methods for \textbf{Office-Home} and \textbf{DomainNet}. In addition, ECA neither has source data nor domain labels, but it even outperforms ConMix that adopts domain labels during training on all datasets. Therefore, ECA outperforms other methods with considerable gains compared with those that have domain labels or source data in prior.



\section{Discussion}

\subsection{Ablation studies}

\begin{figure}[t]
    \centering  
    \includegraphics[width=0.45\textwidth]{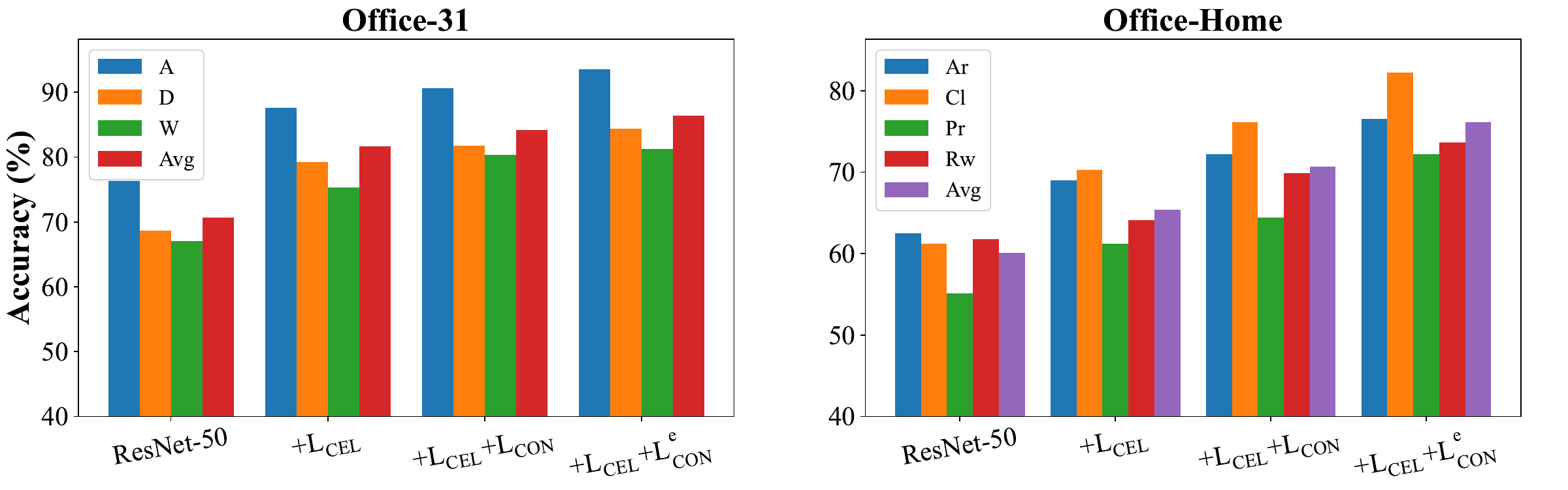}
    \caption{Overall Accuracy (\%) on \textbf{Office-31} and \textbf{OfficeHome} for ablation studies.}
    \label{fig:abla}
    \vspace{-1em}
\end{figure}

\begin{figure}[t]
    \centering  
    \includegraphics[width=0.45\textwidth]{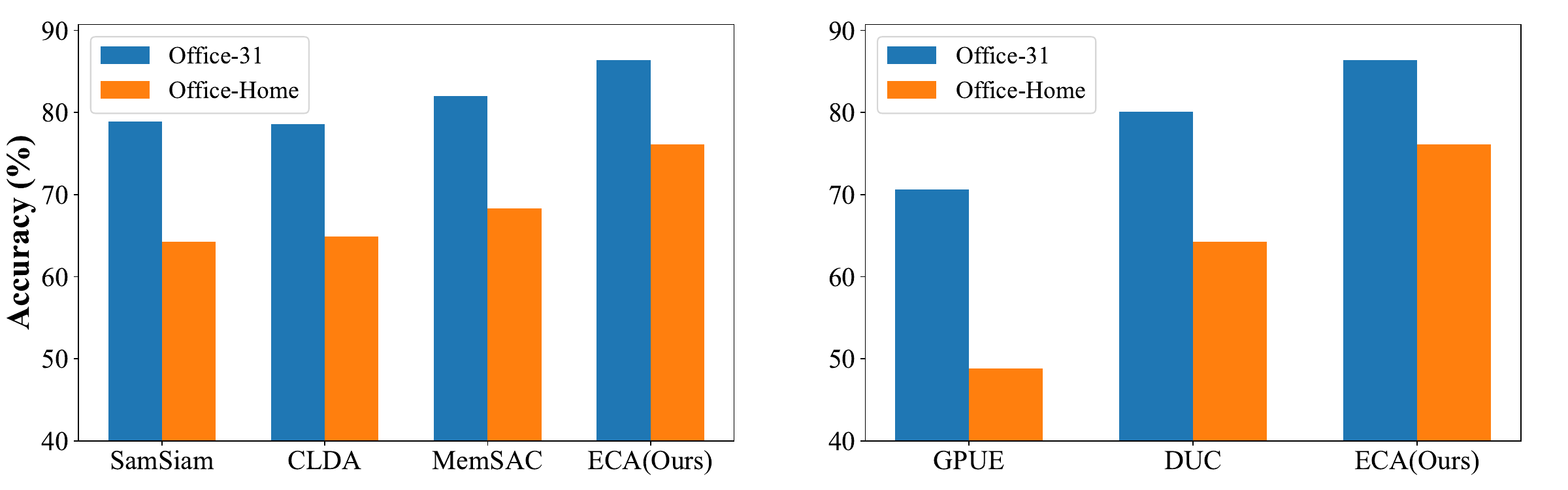}
    \caption{Comparisons among other contrastive learning methods and other uncertainty methods with our  ECA.}
    \label{fig:comparisons}
    \vspace{-1em}
\end{figure}

As listed in Fig. \ref{fig:abla}, $L_{\text{CON}}$ achieves $2.5\%$ and $3.8\%$ improvement for \textbf{Office-31} and \textbf{OfficeHome}, respectively. 
In addition, $L_\text{CON}^e$ further increase $1.1\%$ and $1.4\%$ for average accuracy. More ablation studies and those on \textbf{DomainNet} can be found in our supplementary.
In addition, Table \ref{tab:conf} lists ablation results for different high-quality selection schemes. Therefore, we both utilize the evidential uncertainty and prediction confidence to measure the label reliability simultaneously (see more details in Sec. \ref{sec:CEU}). 
Furthermore, we compare different contrastive learning methods (such as SamSiam \cite{chen2021exploring}, CLDA \cite{singh2021clda}, MemSAC \cite{kalluri2022memsac}) and uncertainty methods (such as GPUE \cite{litrico2023guiding} and DUC \cite{xie2022dirichlet}) in Fig. \ref{fig:comparisons}. Experimental results indicates that the evidential contrastive learning and the calibrated evidential learning in our proposed ECA perform the best among other state-of-the-art methods.

\begin{table}[t]
\centering
\small
\begin{tabular}{cccc}
\hline
\textbf{Method} & \textbf{Office-31}    & \textbf{Office-Home}  &  \textbf{DomainNet}  \\ \hline
only $\eta_u$ & 85.9 & 74.3 & 32.8 \\
only $\eta_c$ & 85.6 & 75.5 & 32.2 \\
both $\eta_u$ and $\eta_c$ & \textbf{86.4} & \textbf{76.1} & \textbf{34.6} \\ \hline

\end{tabular}
\caption{Accuracy on different sample selection schemes.}
\label{tab:conf}
\vspace{-1em}
\end{table}


\begin{table}[t]
\centering
\small

\begin{tabular}{cccccc}
\hline

\textbf{$k$} & 2 & 3 & 4 & 5 & 6\\ \hline

Office-31 &  86.4* & 86.1 & \textbf{86.9} & 85.7 & 86.6  \\
Office-Home  & 74.4 & 76.1* & \textbf{77.0} & 76.1  & 76.6 \\
DomainNet  & 24.2 & 32.9 & 33.3 & \textbf{34.6}* & 33.8 \\\hline

\end{tabular}
\caption{Overall accuracy (\%) on \textbf{Office-31}, \textbf{Office-Home} and \textbf{DomainNet} with different $k$. * denotes reported result for our proposed ECA in Table \ref{tab:office} and Table \ref{tab:domainnet}.}
\label{tab:k}
\vspace{-2em}
\end{table}

\subsection{Do We Need Domain Label or Source Data?}

According to Table \ref{tab:office}, source data is a crucial impact on \textbf{Office-31}, while domain label is not the necessary information during adaption. On the other hand, our proposed ECA achieves highest accuracy compared with those methods with source data or domain label on \textbf{Office-Home}. 
As for \textbf{DomainNet}, source data and domain label are not the necessary conditions. In conclusion, \textbf{we may not need domain label nor source data in SF-BTDA scenarios.}

Furthermore, Table \ref{tab:k} displays overall accuracy on different datasets with different $k$. We observe that choosing $k$ as the number of sub-targets may not leads to the superior performance. While when the $k$ is larger, the accuracy only has slight changes. Therefore, we select the same value of $k$ as the true number of target domains in each dataset to report.


\subsection{Theoretical understanding}

Inspired by the generalized label shift theorem in \cite{tachet2020domain}, in SF-BTDA setting, for any classifier $\hat{Y} = \left(h \circ g \right)\left(X\right)$, the blended target error rate could be formulated as $\left\Vert \frac{1}{K} \sum_j^{K} \epsilon_{T_j}  \right \Vert \leq  BER_{f_s} \left( \hat{Y} \Vert Y \right) + 2\left( c-1 \right)\Delta_{BTCE}\left( \hat{Y} \right)$. $BER_{f_s} \left( \hat{Y} \Vert Y \right)$ is the classification performance only related with the source model. $\Delta_{BTCE}$ measures the conditional distribution discrepancy of each class between the the source and each target. 
In this sense, we only need to minimize the $\Delta_{BTCE}\left( \hat{Y} \right)$. Therefore, the most important objective to improve the quality of pseudo labels, which may lead to category misalignment and negative transfer effects. In addition, Our Supplementary has visualized our theoretical understanding including the values of $\mathcal{H} \Delta \mathcal{H}$-distance (measuring domain shift as the discrepancy between the disagreement of two hypotheses $h, h' \in \mathcal{H} \Delta \mathcal{H}$) and $\lambda$ (the ideal joint hypothesis $h^*$ on both source and target domains).

\subsection{Sensitive Analysis}

There are much less hyper-parameters need to tune.  We display the performance of different $\beta$ in Eq. \eqref{eq:total} in our supplementary. We set $\beta$ as 1.0 for the whole experiments. 
In addition, we will discuss the annealing weight $\gamma$ introduced in Eq. \eqref{eq:evidential} and comparisons between adaptive selection and fixed hyper-parameters for $\eta_c$ and $\eta_u$ in our supplementary.


\subsection{How ECA performs in CTTA scenario?}

In this paper, we focus on SF-BTDA setting, while it is surprising that our proposed ECA also achieves the highest accuracy for Continual Test-Time Adpaptation (CTTA) scenario (See our supplementary). In CTTA, the target data is provided in a sequence and from a continually changing environment. In both of SF-BTDA and CTTA, the adaptation of the target network does not rely on any source data. 

\section{Conclusions}

In this paper, we are the first to propose a new DA settings (\textit{i.e.}, SF-BTDA), which is a more practical and challenging task where we can not access to source domain data while facing mixed multiple target domains without any domain labels in prior. Towards to address this scenario, we propose Evidential Contrastive Alignment (ECA) and conduct a new benchmark for SF-BTDA. ECA effectively addressing two major challenges in SF-BTDA: (1) facing a mixture of multiple target domains, the model is assumed to handle the co-existence of different label shifts in different targets with no prior information;
and (2) as we are only accessed to the source model, precisely alleviating the negative effect from noisy target pseudo labels is difficult because of the significantly enriched styles and textures.  We conduct extensive experiments on three DA benchmarks, and empirical results show that ECA outperforms state-of-the-art SF-STDA methods with considerable gains and achieves comparable results compared with those that have domain labels or source data.

\bibliography{aaai25}

\end{document}